\documentclass[letterpaper]{article} 
\usepackage{aaai25}  
\usepackage{times}  
\usepackage{helvet}  
\usepackage{courier}  
\usepackage[hyphens]{url}  
\usepackage{graphicx} 
\usepackage{natbib}  
\usepackage{caption} 
\usepackage{algorithm}
\usepackage{algorithmic}
\usepackage{amsmath}
\usepackage{amssymb}
\usepackage{booktabs}
\usepackage{array}
\usepackage{multirow}
\usepackage{colortbl}
\usepackage{colortbl}
\usepackage{xcolor}
\definecolor{lightgray}{gray}{0.9} 
\usepackage{subcaption} 
\usepackage{booktabs}
\usepackage{xspace}

\def\eg{\emph{e.g.}\xspace}

\def\etc{\emph{etc.}\xspace}

\usepackage{newfloat}
\usepackage{listings}
\DeclareCaptionStyle{ruled}{labelfont=normalfont,labelsep=colon,strut=off} 
\floatstyle{ruled}
\newfloat{listing}{tb}{lst}{}
\floatname{listing}{Listing}
\title{FOCUS: Towards Universal Foreground Segmentation}
\author {
    Zuyao You\textsuperscript{\rm 1,\rm 2},
    Lingyu Kong\textsuperscript{\rm 1}\equalcontrib,
    Lingchen Meng\textsuperscript{\rm 1,\rm 2}\equalcontrib,
    Zuxuan Wu\textsuperscript{\rm 1,\rm 2\thanks{Corresponding author.}},
}
\affiliations {
\textsuperscript{\rm 1}Shanghai Key Lab of Intell. Info. Processing, School of CS, Fudan University \\
\textsuperscript{\rm 2}Shanghai Collaborative Innovation Center of Intelligent Visual Computing\\
    \{zyyou23, lykong22\}@m.fudan.edu.cn, \{lcmeng20, zxwu\}@fudan.edu.cn \\
    \bigskip
    \textbf{\textcolor{magenta}{https://geshang777.github.io/focus.github.io/}}

}

\usepackage{bibentry}

\begin{document}

\maketitle
\begin{abstract}

Foreground segmentation is a fundamental task in computer vision, encompassing various subdivision tasks. Previous research has typically designed task-specific architectures for each task, leading to a lack of unification. Moreover, they primarily focus on \textbf{recognizing foreground objects} without effectively \textbf{distinguishing them from the background}. In this paper, we emphasize the importance of the background and its relationship with the foreground. We introduce \textbf{FOCUS}, the \textbf{F}oreground \textbf{O}bje\textbf{C}ts \textbf{U}niversal \textbf{S}egmentation framework that can handle multiple foreground tasks. We develop a multi-scale semantic network using the edge information of objects to enhance image features. To achieve boundary-aware segmentation, we propose a novel distillation method, integrating the contrastive learning strategy to refine the prediction mask in multi-modal feature space. We conduct extensive experiments on a total of \textbf{13 datasets} across \textbf{5 tasks}, and the results demonstrate that FOCUS consistently outperforms the state-of-the-art task-specific models on most metrics.

\end{abstract}


\section{Introduction}

\begin{figure}[t]
\centering
\includegraphics[width=\columnwidth]{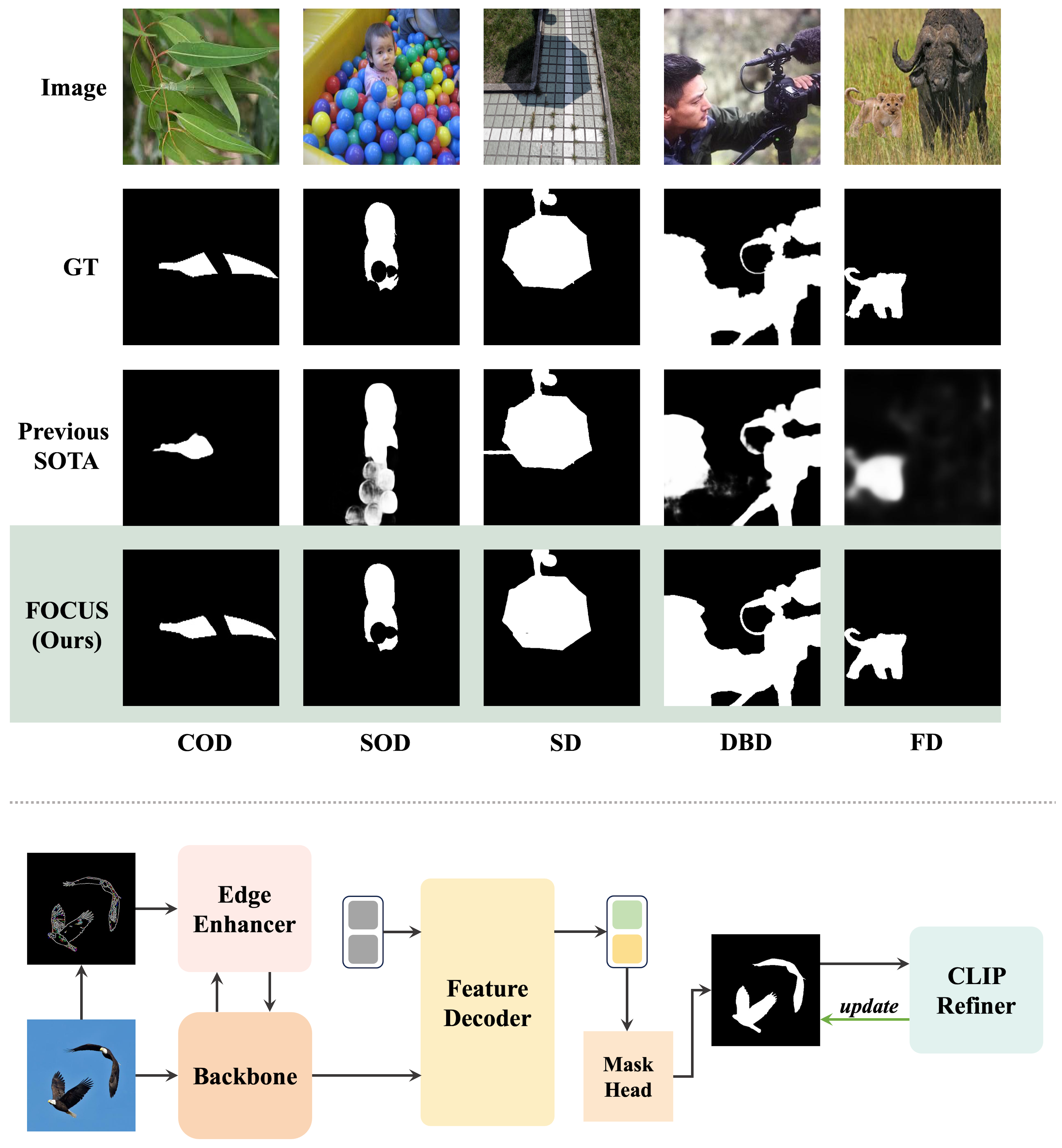} %
\caption{With one unified architecture, FOCUS can handle various foreground segmentation tasks. Our proposed method can generate boundary-aware masks that are smoother and more detailed than the previous state-of-the-art task-specific models. Zoom in for more details.}
\label{fig1}
\end{figure}

Foreground segmentation is a fundamental task in computer vision where the primary goal is to delineate the prominent objects (foreground) from the rest of the image (background), typically referring to salient object detection (SOD) and camouflaged object detection (COD) \cite{ZoomNet-CVPR2022, ZoomNeXt}. In this paper, the concept of foreground segmentation can be extended to delineating objects that interest you most in the image, where the primary goal is to obtain the Mask of Interest (MoI), \eg, MoI should denote the mask of the camouflaged object in COD. According to this definition, tasks such as shadow detection (SD), defocus blur detection (DBD), forgery detection (FD), \etc belong to the category of foreground segmentation, too.

Currently, in the field of generic segmentation, \eg instance segmentation, semantic segmentation, and panoptic segmentation, \etc, there are already many sophisticated models \cite{Kirillov_2023_ICCV,cheng2022masked,jain2023oneformer,ding2023mose,Ding_2023_ICCV}. However, these models often lack targeted training for specific foreground segmentation tasks. For instance,  in the COD task, SAM struggles to distinguish camouflaged objects from the background \cite{hu2024relax}. Furthermore, without prompt-guided methods, most traditional segmentation algorithms generate multiple masks for one image at the same time \cite{cheng2021per,cheng2022masked,jain2023oneformer}, but users do not require such many masks in many real-world scenarios, \eg image background removal, MoI is all they need. While foreground segmentation typically produces a single or specific type of mask, making it more in line with user needs.

However, as mentioned earlier, when the concept of foreground is generalized as MoI, the scope of foreground segmentation tasks is very broad. Currently, there is a lack of an excellent and universal framework that can handle all foreground segmentation tasks. Most foreground segmentation models are task-specific \cite{wang2022objectformer,zhao2021defocus,zhu2021mitigating,zheng2024birefnet,xie2022pyramid,wang2022m2tr} . Some models \cite{ZoomNeXt, ZoomNet-CVPR2022} achieve universality in SOD and COD tasks, but given the similarity between COD and SOD tasks, they will not be discussed as universal models here. To the best of our knowledge, the work most closely related to ours is \cite{liu2023explicit}. However, it still significantly lags behind task-specific models after fine-tuning in the subdivision tasks.

\begin{figure*}[t]
\centering
\includegraphics[width=1.0\textwidth]{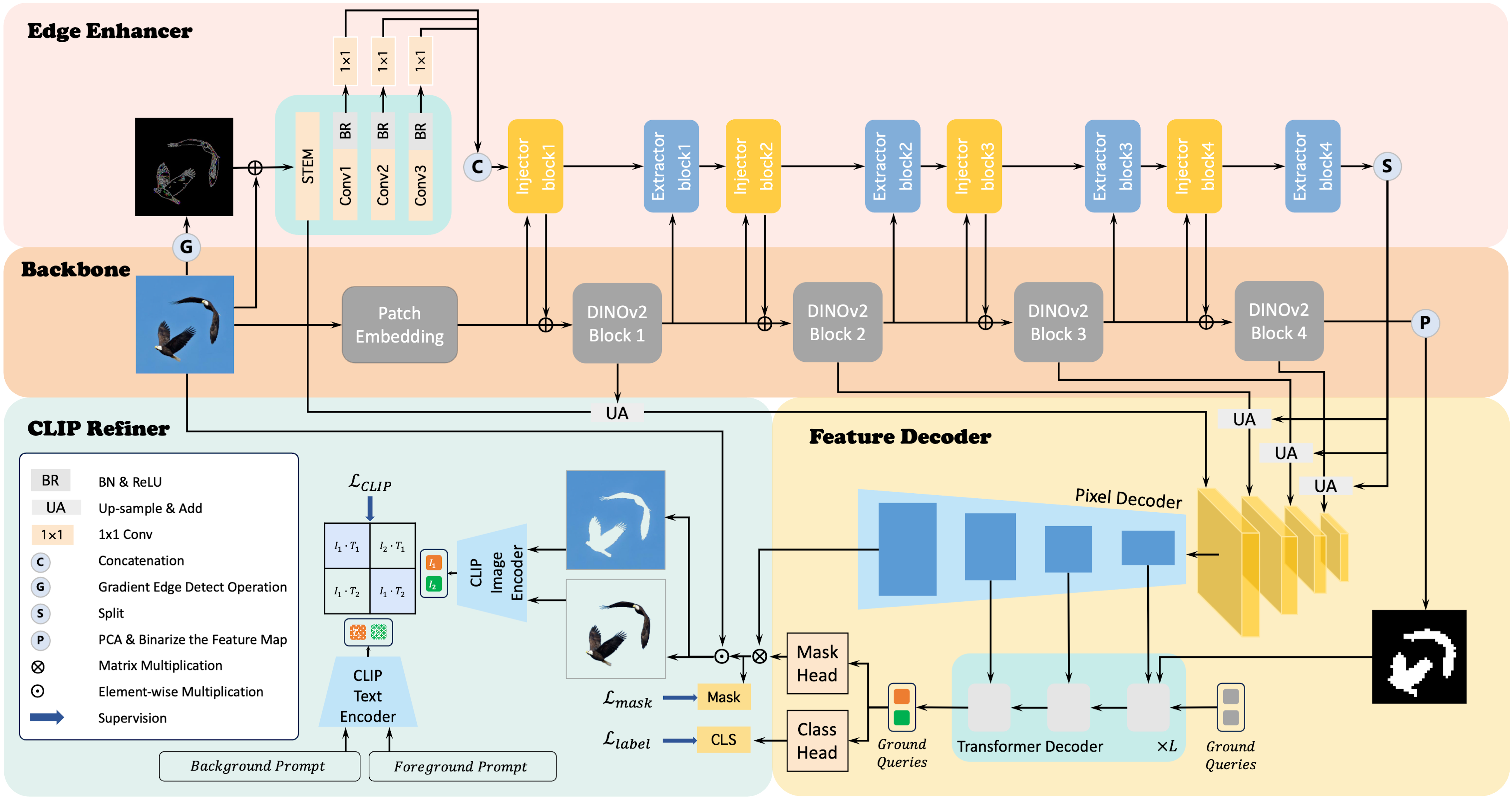} 
\caption{An overview of our proposed FOCUS, a multi-scale and multi-modal semantic framework for universal foreground segmentation, mainly includes the backbone, edge enhancer, feature decoder, and CLIP refiner. Refer to the main text for details.}
\label{framework}
\end{figure*}

Besides, previous foreground segmentation models primarily focused on recognizing the foreground objects without effectively distinguishing them from the background, neglecting the background and the relationship between the background and the foreground. In fact, background information plays a critical role in computer vision tasks \cite{li2023mask, meng2024learning}.
Foreground segmentation inherently involves distinguishing the foreground from the background, making both elements and their relationship vital. However, current approaches fail to address the background segmentation separately. Consequently, this oversight impacts the overall performance of foreground segmentation.

The issues above can be summarized as follows: \textbf{(1)}How to generally represent the foreground and background of different foreground segmentation tasks? \textbf{(2)}How to fully utilize the background information of an image to optimize prediction results? In this paper, we introduce \textbf{FOCUS}, a unified multi-modal approach to tackle multiple subdivision tasks of foreground segmentation.

To universally represent the foreground and background, we borrow the object queries concept from DETR \cite{carion2020end} by introducing ground queries. We apply the multi-scale strategy \cite{cheng2022masked} to extract image features to feed the transformer decoder, using masked attention to enable the ground queries to focus on relevant features corresponding to foreground and background. We utilize the feature map obtained from the backbone to initialize the masked attention, which can serve as a localization prior. During this process, the ground queries adapt to learn the features relevant to the context of different tasks, making them universal features.

To fully leverage the background information in images, we employ contrastive learning strategies. We propose the CLIP refiner, using the powerful multi-modal learning ability from CLIP \cite{radford2021learning} to correct the masks generated by previous modules. We fuse the mask and image and align the fused image and its corresponding text in multi-modal feature space to refine the masks. This not only refines the edges of the mask but also accentuates the distinction between foreground and background. We treat foreground segmentation and background segmentation as two independent tasks, and in the inference stage, the probability map of both foreground and background will jointly determine the boundary of MoI.


We conduct detailed experiments on 13 datasets across five foreground segmentation tasks and achieve or exceed state-of-the-art on most provided metrics. Fig.~\ref{fig1} shows the outstanding performance of our proposed FOCUS on different sub-tasks of the foreground segmentation. 

Our contributions can be summarized as follows:
\begin{itemize}
    \item We propose a unified framework for foreground segmentation tasks, including SOD, COD, SD, DBD, and FD;
\end{itemize}
\begin{itemize}
    \item We propose a novel module, using the contrastive learning strategy to utilize the background information to refine the mask while widening the distance between the foreground and the background; 
\end{itemize}
\begin{itemize}
    \item We conduct extensive experiments on multiple datasets across multiple tasks, and results demonstrate that our method achieves state-of-the-art performance.
\end{itemize}

\section{Related Work}

\subsection{Foreground Segmentation}
As mentioned earlier, several tasks are crucial in foreground segmentation, including salient object detection(SOD), camouflaged object detection (COD), shadow detection (SD), defocus blur detection (DBD), and forgery detection (FD). SOD aims at segmenting the most visually attractive objects from the input images. COD focuses on disguised objects that blend seamlessly into their surroundings, \eg mimetic animals and body paintings. SD aims to segment shadow regions from natural scenes. DBD aims at separating in-focus and out-of-focus regions,  which is caused by the different focal lengths of the cameras, slightly different from SOD. The goal of FD is to identify altered or manipulated areas in images, typically involving addition, replacement, or deletion. Previous models normally designed architectures for specific foreground segmentation task \cite{wang2022objectformer,zhao2021defocus,zhu2021mitigating,zheng2024birefnet,xie2022pyramid}, and currently, there is a lack of effective methods to handle this foreground segmentation tasks universally. 

\subsection{Universal Segmentation}
Universal segmentation has emerged as a significant trend in computer vision. It aims to unify various segmentation tasks within a single framework. 
This trend started with efforts to unify semantic and instance segmentation through panoptic segmentation \cite{pq} and has since expanded to include a broader range of tasks.
Recent works have shifted towards designing universal segmentation models with generalization ability and versatility. 
Mask2Former \cite{cheng2022masked} utilizes a masked-attention mechanism to unify instance, semantic and panoptic segmentation. 
OneFormer \cite{jain2023oneformer} further improves Mask2Former with a multi-task train-once design. 
More recent approaches like SAM \cite{Kirillov_2023_ICCV} push the boundaries of universal segmentation with the ability of zero-shot segmentation.
In the field of foreground segmentation, the unified architecture most related to ours is EVP \cite{liu2023explicit}. EVP freezes a pre-trained model and then learns task-specific knowledge using an adapter structure, but its performance falls behind task-specific models. In this work, we aim to find a more effective way to unify the foreground segmentation tasks using one single architecture.

\section{Methods}

\subsection{Unified Architecture}
Previously, there was a lack of unified architecture for handling all foreground segmentation subdivision tasks. Given an image from different foreground segmentation tasks, our goal is to use a unified architecture to predict the corresponding MoI in the task context. The problem can be defined by:
\[ U(I, T_i) = MoI \]
 $T_i$ refers to different foreground segmentation tasks, $\forall T_i \in \{T_1, \ldots, T_n\}$ the unified framework $U$ should infer the corresponding $MoI$ from the images $I$.

We propose FOCUS, a unified architecture that can handle multiple foreground segmentation tasks. We borrow the concept of object queries from \cite{carion2020end} and introduce the ground queries (\(\mathbf{GQ}\)) here. \(\mathbf{GQ}\) are two distinct tensors, designated as the foreground query and background query, we aim to only use these two learned tensors to respectively embed and represent the foreground and the background within the image based on the context of the task. Fig.~\ref{framework} provides an overview of our approach FOCUS. After obtaining multi-scale edge-enhanced features from the backbone and the edge enhancer, the pixel decoder will generate pixel-level output and these pixel-level features will be fed into the transformer decoder with \(\mathbf{GQ}\), where \(\mathbf{GQ}\) updated by masked attention \cite{cheng2022masked} to get ground-centric output. It can be formulated as:

\[
\mathbf{X}_l = \text{softmax}(\mathcal{M}_{l-1} + \mathbf{GQ}_l \mathbf{K}_l^\top) \mathbf{V}_l + \mathbf{X}_{l-1},
\]

Here, \(\mathbf{K}_l,\mathbf{V}_l\ \in \mathbb{R}^{H_l W_l\times C}\) denotes the linearly transformed \(C\)-dimensional image feature from \(\mathbf{l}_{th}\) block of pixel decoder, \(\mathbf{X}_l \in \mathbb{R}^{2 \times C}\) refers to the query feature from the  \(\mathbf{l}_{th}\) transformer decoder block and \(\mathbf{X}_0 \) is initialized by the input query feature of transformer decoder. \(\mathbf{GQ}_l \in \mathbb{R}^{2 \times C}\) is the \(\mathbf{l}_{th}\) ground queries, and \(\mathcal{M}_{l-1}\) is defined by:

\[
\mathcal{M}_{l-1}(x, y) = 
\begin{cases} 
0 & \text{if } 
\mathbf{M}_{l-1}(x, y) = 1 \\
-\infty & \text{otherwise}
\end{cases},
\]

 \(\mathbf{M}_{l-1} \in \{0, 1\}^{2 \times H_l W_l}\) is obtained by decoding \(\mathbf{GQ}_
 {l-1} \) and binarizing, with dimensions resizing consistent with \(\mathbf{K}_l\). DINOv2 \cite{oquab2023dinov2} is a recently proposed model designed for visual representation learning. The visualization of its feature map indicates that DINOv2 has already focused on the prominent objects in the image without supervision, showing richer semantics compared to other foundation models \cite{wang2022pvt, meng2022adavit}. Therefore, we choose DINOv2 as the backbone for FOCUS, PCA and binarize the feature map of its last block to initialize the attention mask \(\mathcal{M}_{0}\).  \(\mathcal{M}_{0}\) is formulated as:

 \[
\mathcal{M}_{0}(x, y) = 
\begin{cases} 
0 & \text{if } \mathbf{F}_{DINOv2}(x, y) = 1 \\
-\infty & \text{otherwise}
\end{cases}.
\]
 
 Here,  \(\mathbf{F}_{DINOv2}\) refers to the binary feature map from the last backbone block. It is resized to the same resolution of \(\mathbf{K}_1\). The adoption of the new initialization method can leverage the localization prior knowledge learned by the DINOv2 on large-scale data.
 
We use two multi-layer perceptrons, designated as mask head and class head, to decode ground queries and generate mask and class predictions for both the foreground and background. During the inference stage, the foreground and background probability distributions are combined to predict the final MoI.
 
\subsection{Edge Enhancer}

 
 In order to utilize the edge information of the object, we propose the edge enhancer, an effective module that uses foreground object edge information to correct the image features obtained by the backbone. 
 
Inspired by the recent study that shows convolutions can help transformer understand local spatial information \cite{chen2022vision,wang2022pvt}, we use ResNet50 \cite{he2016deep} to extract edge features from the image. We convert the image into grayscale to reduce the confusion caused by color, apply Gaussian smoothing \cite{davies2004machine} to reduce noise, and then use an edge detector \cite{canny1986computational} to obtain a gradient map and overlay it on the original image. As shown in Fig.~\ref{framework}, the ResNet can be divided into the STEM and the rest, the STEM serves as the initial feature extractor, comprising a series of convolution, batch normalization, and ReLU activation layers. The output of the rest convolution blocks will be flattened and projected into the same dimension \(D\) by 1×1 convolutions and concatenated to obtain feature pyramid \(F_{\text{edge}}^1 \in \mathbb{R}^{(\frac{HW}{8^2} + \frac{HW}{16^2} + \frac{HW}{32^2}) \times D}\), \(H\) and \(W\) represent the resolution of the input image. Then, we follow ViT-Adapter \cite{chen2022vision}, using the structure of the injector-extractor based on cross attention to fuse the image features from the backbone and ResNet. The injector can be formulated as:

\[\hat{F}_{\text{DINOv2}}^i = F_{\text{DINOv2}}^i + \gamma^i \text{MSDA}(F_{\text{DINOv2}}^i, F_{\text{edge}}^i),\]

MSDA refers to multi-scale deformable attention \cite{zhu2020deformable}, which takes the normalized backbone feature \(F_{\text{DINOv2}}^i \in \mathbb{R}^{\frac{HW}{16^2} \times D}\) as the query, and the normalized edge feature \(F_{\text{edge}}^i \in \mathbb{R}^{(\frac{HW}{8^2} + \frac{HW}{16^2} + \frac{HW}{32^2}) \times D}\) as the key and value. \(\gamma^i\) is a learnable parameter for balancing the backbone feature and the fused feature. Similarly, the extractor can be formulated as:


\[\hat{F}_{\text{edge}}^i = F_{\text{edge}}^i + \text{ConvFFN}(\text{MSDA}(F_{\text{edge}}^i, F_{\text{DINOv2}}^{i+1})).\]

It is another multi-scale deformable attention like injector while taking the normalized edge feature $F_{\text{edge}}^i \in \mathbb{R}^{(\frac{HW}{8^2} + \frac{HW}{16^2} + \frac{HW}{32^2}) \times D}$ as the query, and the output feature $F_{\text{DINOv2}}^{i+1} \in \mathbb{R}^{\frac{HW}{16^2} \times D}$ as the key and value. \(\text{ConvFFN}\) refers to the structure with two fully connected layers and a depth-wise separable convolution layer. The \(\hat{F}_{\text{edge}}^i\)
will serve as the input for the next injector. We upscale the output from different blocks of backbone to resolutions of 1/4, 1/8, 1/16, and 1/32. Besides, we split the output of the last extractor, and restore them to their original size. Then we add the up-scaled backbone features with the corresponding split output from extractor and output from STEM to get the edge-enhanced multi-scale image features. These features will be fed into the pixel decoder, another module based on multi-scale deformable attention, for dense pixel-level predictions. 

\begin{table*}[htbp] 
    \centering

    \setlength{\tabcolsep}{1mm} 
    \scriptsize
    \begin{tabular}{c|*{4}{c}|*{4}{c}|*{4}{c}|*{4}{c}}
        \toprule
        & \multicolumn{4}{c}{CAMO(250)} 
        & \multicolumn{4}{c}{COD10K(2,026)} 
        & \multicolumn{4}{c}{CHAMELEON(76)}  
        & \multicolumn{4}{c}{NC4K(4,121)}  \\
        & \(\mathcal{S}_m \uparrow\)  & \({E_{\xi}} \uparrow\) & \({F_{\beta}^w} \uparrow\) &  \(MAE \downarrow\) 
        & \(\mathcal{S}_m \uparrow\)  & \({E_{\xi}} \uparrow\) & \({F_{\beta}^w} \uparrow\) &  \(MAE \downarrow\)  
        & \(\mathcal{S}_m \uparrow\)  & \({E_{\xi}} \uparrow\) & \({F_{\beta}^w} \uparrow\) &  \(MAE \downarrow\)   
        & \(\mathcal{S}_m \uparrow\)  & \({E_{\xi}} \uparrow\) & \({F_{\beta}^w} \uparrow\) &  \(MAE \downarrow\)  \\
        \midrule
        SINet$_{20}$    &.751 &.771 &.606 &.100   & .771 &.806 &.551 &.051   & .869 &.891 &.740 &.044   &.808 &.871 &.723 &.058 \\
        PFNet$_{22}$    &.782 &.852 &.695 &.085   &.800 &.868 &.660 &.040   &.882 &.942 &.810 &.033   &.829 &.898 &.745 &.053 \\
        ZoomNet$_{22}$    &.820 &.892 &.752 &.066   & .838 &.911 &.729 &.029   & .902 &.958 &.845 &.023   &.853 &.912 &.784 &.043 \\
        BSA-Net$_{22}$    &.794 &.867 &.717 &.079   & .818 &.901 &.699 &.034   & .895 &.957 &.841 &.027   &.842 &.907 &.771 &.048 \\
        FSPNet$_{23}$    &.856 &.899 &.799 &.050   & .851 &.895 &.735 &.026   & .908 &\underline{.965} &.851 &.023   &.879 &.915 &.816 &.035 \\
        ZoomNeXt$_{24}$  &.889 &.945 &.857 &.041   & .898 &.956 &.827 &.018   & \underline{.924} &\textbf{.975} &.885 &.018   &.903 &.951 &.863 &.028 \\
        BiRefNet$_{24}$  &\underline{.904} &\underline{.954} &.\underline{890} &\underline{.030}   & \textbf{.912} &\underline{.960} &\underline{.874} &\underline{.014}   & \textbf{.932} &- &\textbf{.915} &\textbf{.015}   &\underline{.914} &\underline{.953} &\underline{.894} &\underline{.023} \\
        \rowcolor{lightgray}
        \rowcolor{lightgray}
        \rowcolor{lightgray}
        \textbf{FOCUS(Ours)}   &\textbf{.912} &\textbf{.963} &\textbf{.904} &\textbf{.025}   & \underline{.910} &\textbf{.974} &\textbf{.883} &\textbf{.013}  &.922 &\textbf{.975} &\underline{.908} &\underline{.017}   &\textbf{.915} &\textbf{.964} &\textbf{.906} &\textbf{.020} \\
        \bottomrule
    \end{tabular}
    \caption{Comparison of FOCUS with recent state-of-the-art COD methods.}
    \label{tab:COD}
\end{table*}

\begin{table*}[htbp] 
    \centering

    \setlength{\tabcolsep}{1.0mm} 
    \scriptsize
    \begin{tabular}{c|*{3}{c}|*{3}{c}|*{3}{c}|*{3}{c}|*{3}{c}}
        \toprule
        & \multicolumn{3}{c}{DUTS-TE(5,019)} 
                               & \multicolumn{3}{c}{DUT-OMRON(5,618)} 
                               & \multicolumn{3}{c}{HKU-IS(4,447)} 
                               & \multicolumn{3}{c}{ECSSD(1,000)}  
                               & \multicolumn{3}{c}{PACAL-S(850)}  \\
                               & \(\mathcal{S}_m \uparrow\)  & \({E_{\xi}} \uparrow\)  &  \(MAE \downarrow\) 
                               & \(\mathcal{S}_m \uparrow\)  & \({E_{\xi}} \uparrow\)  &  \(MAE \downarrow\)  
                               & \(\mathcal{S}_m \uparrow\)  & \({E_{\xi}} \uparrow\)  &  \(MAE \downarrow\)  
                               & \(\mathcal{S}_m \uparrow\)  & \({E_{\xi}} \uparrow\)  &  \(MAE \downarrow\)  
                               & \(\mathcal{S}_m \uparrow\)  & \({E_{\xi}} \uparrow\)  &  \(MAE \downarrow\)  \\
        \midrule
        VST$_{21}$    &.896 &.892 &.037   & .850 &.861 &.058   &.928 &.953 &.029   &.932 &.918 &.033   &.865 &.837 &.061 \\
        BBRF$_{21}$     &.908 &.927 &.025   &.855 &.887 &\textbf{.042}   &\textbf{.935} &.965 &\underline{.020}  &.939 &.934 &\underline{.022}   &.871 &.867 &\underline{.049} \\
        EVPv1$_{23}$     &.913 &.947 &.026 &\underline{.862} &.894 &.046 &.931 &.961  &.024 &.935 &\underline{.957}  &.027 &.878 &\underline{.917} &.054 \\
        EVPv2$_{23}$     &.915 &\underline{.948}  &.027 &\underline{.862} &\underline{.895} &.047 &.932 &.963 &.023 &.935 &\underline{.957}  &.028 &\underline{.879} &\underline{.917} &.053 \\

        MENet$_{23}$   &.905 &.937 &.028   &.850 &.891 &.045 &.927 &\underline{.966} &.023 &.928 &.954 &.030
        &872 &.913 &.054\\

        SelfReformer$_{23}$     &\underline{.921} &.924 &\underline{.024}   &.859 &.884 &\underline{.043}  &\underline{.934} &.961 &.023   &\underline{.941} &.935 &.025   &.877 &.874 &\underline{.049} \\

        \rowcolor{lightgray}
        \rowcolor{lightgray}
        \rowcolor{lightgray}
        \textbf{FOCUS(Ours)}       &\textbf{.929} &\textbf{.965} &\textbf{.019}   &\textbf{.868} &\textbf{.900} &.045  &\textbf{.935} &\textbf{.974} &\textbf{.018}   &\textbf{.943} &\textbf{.971} &\textbf{.018}   &\textbf{.898} &\textbf{.942} &\textbf{.036} \\
        \bottomrule
    \end{tabular}
    \caption{Comparison of FOCUS with recent state-of-the-art SOD methods.}
    \label{tab:SOD}
\end{table*}

\subsection{CLIP Refiner}


Since the proposal of CLIP, there have been many works using CLIP for segmentation \cite{xu2022groupvit,li2022language,wang2022cris,liang2023open}, which have proven that CLIP is effective not only at the image level but also at the pixel level. In this paper, we propose CLIP refiner, which uses the powerful multi-modal ability of CLIP to correct the masks of foreground and background. 

Specifically, we decode the ground queries to obtain the masks of the foreground and background, resize them, and overlay them on the image. We use the prompts ``It's an image of salient objects without background.'' and ``It's an image of background with salient objects removed.'' to represent foreground and background, respectively. Note that the text can be adjusted according to the task. For example, in shadow detection, prompts can be replaced with ``it's an image of shadow without background.'' and ``it's an image of background without shadow.'' to extend CLIP refiner to other foreground segmentation tasks. We borrow the image encoder and text encoder from CLIP to encode the image and text separately. Then, we calculate the contrastive loss (\(\mathcal{L}_{\text{clip}}\)) between the mask-fused-image and text features. 

\begin{align*}
\mathcal{L}_{\text{i2t}} = -\frac{1}{2} \bigg[ & \log \frac{\exp(I_f \cdot T_f / \tau)}{\exp(I_f \cdot T_f / \tau) + \exp(I_f \cdot T_b / \tau)} \\
+ & \log \frac{\exp(I_b \cdot T_b / \tau)}{\exp(I_b \cdot T_b / \tau) + \exp(I_b \cdot {T_f} / \tau)} \bigg],
\end{align*}

\begin{align*}
\mathcal{L}_{\text{t2i}} = -\frac{1}{2} \bigg[ & \log \frac{\exp(T_f \cdot {I_f} / \tau)}{\exp({T_f} \cdot {I_f} / \tau) + \exp({T_f} \cdot {I_b} / \tau)} \\
+ & \log \frac{\exp({T_b} \cdot {I_b} / \tau)}{\exp({T_b} \cdot {I_b} / \tau) + \exp({T_b} \cdot {I_f} / \tau)} \bigg],
\end{align*}

\[
\mathcal{L}_{\text{clip}} = \frac{1}{2} (\mathcal{L}_{\text{i2t}} + \mathcal{L}_{\text{t2i}}).
\]

Here $I_{\text{f}}, I_{\text{b}}, T_{\text{f}}, I_{\text{b}} \in \mathbb{R}^{2 \times S}$ denotes the \(S\)-dimensional image feature and text feature of foreground and background obtained by CLIP, \(\tau\) is temperature parameter used to control the smoothness of the softmax function. The CLIP refiner iteratively refines the edges of masks generated by the preceding module, ensuring that only the appropriate pixels are included in the foreground or background. This process aligns the mask-fused image more closely with the corresponding text in the feature space while distancing it from the mismatched one. This not only makes the mask edges more accurate but also widens the gap between the foreground and background. The CLIP refiner is only used to distill knowledge from CLIP and will be discarded during the inference stage. Additionally, we keep the image and text encoders entirely frozen to fully leverage the multi-modal capabilities of CLIP without the potential performance degradation that might arise from fine-tuning.

\begin{figure*}[h]
\centering
\includegraphics[width=0.96\textwidth]{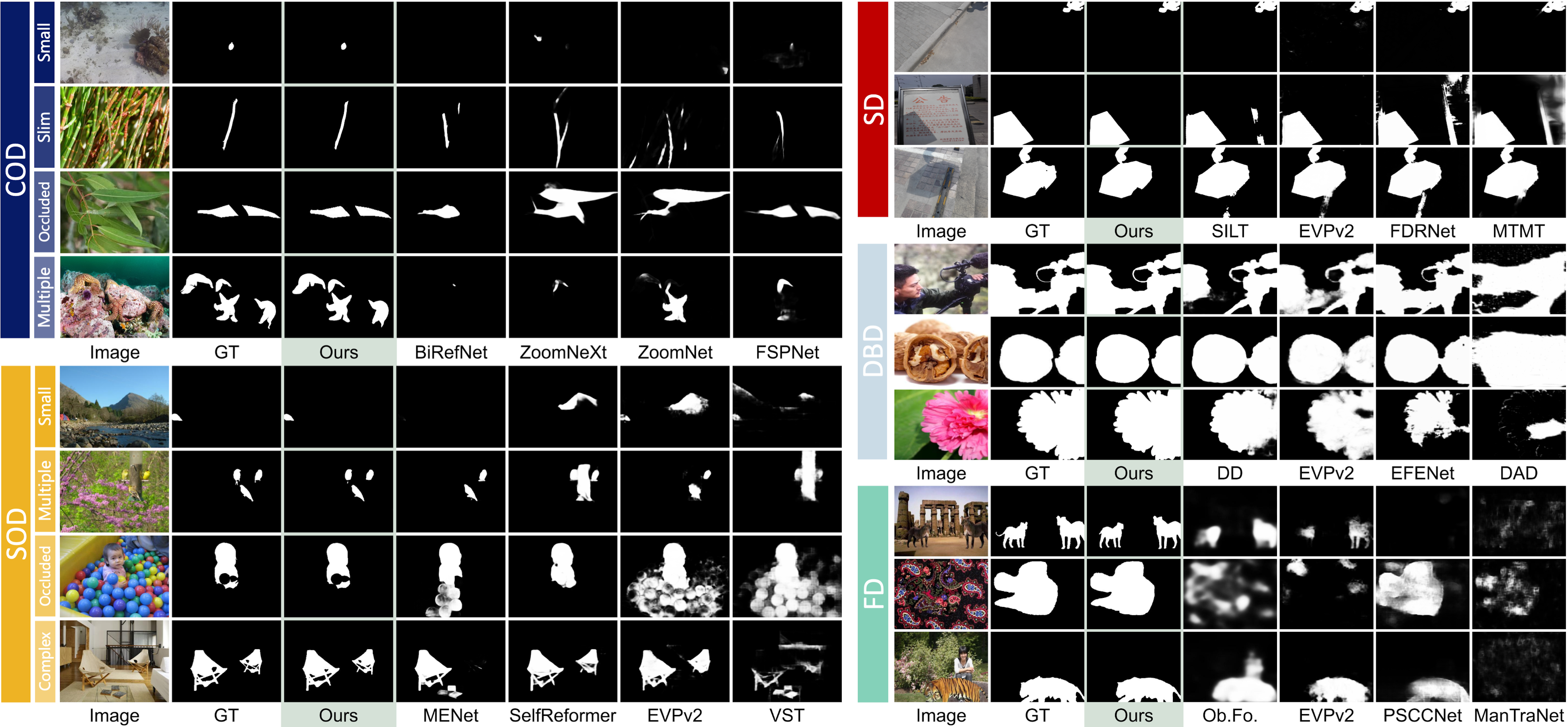} 
\caption{Qualitative comparison of FOCUS and previous methods on COD, SOD, SD, DBD, and FD. Zoom in for more details. }
\label{qualitative1}
\end{figure*}

\begin{table*}[h!]
    \centering

    \begin{subtable}{0.32\textwidth}
        \centering
        \setlength{\tabcolsep}{1.5mm} 
        \scriptsize
        
        \label{tab:SD}
        \begin{tabular}{c|c}
            \toprule
             & ISTD(540) \\
            & \(BER \downarrow\) \\
            \midrule
            BDRAR$_{18}$  & 2.69 \\
            DSD$_{19}$  & 2.17 \\
            MTMT$_{20}$  & 1.72 \\
            FDRNet$_{21}$  & 1.55 \\
            EVPv1$_{23}$  & 1.35 \\
            EVPv2$_{23}$  & 1.35 \\
            SILT$_{23}$  &\underline{1.11} \\
            \rowcolor{lightgray}
            \textbf{FOCUS(Ours)}  & \textbf{0.98} \\
            \bottomrule
            
        \end{tabular}
        \caption{SD}
    \end{subtable}%
    \hfill
    \begin{subtable}{0.32\textwidth}
        \centering
        \setlength{\tabcolsep}{1.5mm} 
        \scriptsize
        
        \label{tab:DBD}
        \begin{tabular}{c|cc|cc}
            \toprule
             & \multicolumn{2}{c}{DUT(500)} & \multicolumn{2}{c}{CUHK(100)} \\
            & \({F_{\beta}} \uparrow\) & \(MAE \downarrow\) & \({F_{\beta}} \uparrow\) & \(MAE \downarrow\) \\
            \midrule
            DeFusionNet$_{20}$ & .823 & .118 & .818 & .117 \\
            CENet$_{19}$ & .817 & .135 & .906 & .059 \\
            DAD$_{21}$ & .794 & .153 & .884 & .079 \\
            EFENet$_{21}$ & .854 & .094 & .914 & .053 \\
            DD$_{20}$ & \underline{.891} & .073 & .927 & \underline{.042} \\
            EVPv1$_{23}$ & .890 & \underline{.068} & .928 & .045 \\
            EVPv2$_{23}$ & .887 & .070 & \underline{.932} & \underline{.042} \\
            \rowcolor{lightgray}
            \textbf{FOCUS(Ours)} & \textbf{.912} & \textbf{.048} & \textbf{.934} & \textbf{.036} \\
            \bottomrule
            
        \end{tabular}
        \caption{DBD}
    \end{subtable}%
    \hfill
    \begin{subtable}{0.32\textwidth}
        \centering
        \setlength{\tabcolsep}{1.5mm} 
        \scriptsize
        
        \label{tab:FD}
        \begin{tabular}{c|cc}
            \toprule
             & \multicolumn{2}{c}{CASIA-1.0(921)} \\
            & \({F1} \uparrow\) & \(AUC \uparrow\) \\
            \midrule
            ManTra$_{19}$ & - & .817 \\
            SPAN$_{20}$ & .382 & .838 \\
            PSCCNet$_{22}$ & .554 & .875 \\
            TransForensics$_{21}$ & .627 & .837 \\
            EVPv1$_{23}$ & .636 & .862 \\
            EVPv2$_{23}$ & \underline{.654} & .876 \\
            ObjectFormer$_{22}$ & .579 & \underline{.882} \\
            \rowcolor{lightgray}
            \textbf{FOCUS(Ours)} & \textbf{.892} & \textbf{.940} \\
            \bottomrule
            
        \end{tabular}
        \caption{FD}
    \end{subtable}
    \caption{Comparison of FOCUS with recent state-of-the-art SD, DBD, and FD methods.}
    \label{tab:three_tables}
\end{table*}

\subsection{Training Objectives}

In order to perform foreground and background segmentation jointly, we convert the foreground segmentation dataset into binary form, with the white areas representing the foreground ground truth and the black areas representing the background ground truth. Following  \cite{cheng2022masked} , we use the combination of binary cross entropy (\(\mathcal{L}_{\text{bce}}\)) and dice loss (\(\mathcal{L}_{\text{dice}}\)) as the loss of the mask, where:

\[\mathcal{L}_{\text{mask}} = \mathcal{L}_{\text{bce}}+\mathcal{L}_{\text{dice}}\]

Recent study \cite{li2023mask} shows that parallel execution of object detection and segmentation can benefit each other. In this paper, we use the rectangular boundary of the ground truth mask as the ground truth bounding box to perform object detection. We use combination of the L1 Regression Loss (\(\mathcal{L}_{\text{L1}}\)) and generalized IoU loss (\(\mathcal{L}_{\text{gIoU}}\)) as the loss for \(\mathcal{L}_{\text{bbox}}\) , which can be formulated as:

\[\mathcal{L}_{\text{bbox}} = \alpha\mathcal{L}_{\text{L1}}+\beta\mathcal{L}_{\text{gIoU}}\]

\(\alpha\) and \(\beta\) are set to 5.0 and 2.0 respectively. We use the standard cross entropy loss as the \(\mathcal{L}_{\text{label}}\). The final training objective is defined as follows:

\begin{align*}
\mathcal{L} = \mathcal{\lambda}_{\text{clip}}\mathcal{L}_{\text{clip}}+\mathcal{\lambda}_{\text{label}}\mathcal{L}_{\text{label}}+\mathcal{\lambda}_{\text{mask}}\mathcal{L}_{\text{mask}}+\mathcal{\lambda}_{\text{bbox}}\mathcal{L}_{\text{bbox}}
\end{align*}

here, \(\mathcal{\lambda}_{\text{clip}}\), \(\mathcal{\lambda}_{\text{label}}\), \(\mathcal{\lambda}_{\text{mask}}\), \(\mathcal{\lambda}_{\text{bbox}}\) refer to the weight of corresponding loss, set to 1.0, 1.0, 5.0, 1.0 respectively. To find the allocation with the lowest cost, we use Hungarian matching \cite{carion2020end,cheng2021per} between the prediction and the ground truth.

\section{Experiments}

\subsection{Datasets and Evaluation Metrics}
For COD, we follow \cite{fan2021concealed,zheng2024birefnet}, training FOCUS on the combination of CAMO-TR \cite{le2019anabranch} and COD10K-TR \cite{fan2020camouflaged} and evaluating on CAMO-TE, COD10K-TE, CHAMELEON \cite{skurowski2018animal} and NC4K \cite{lv2021simultaneously}. We use S-measure (\(\mathcal{S}_m\)), E-measure (\({E_{\xi}} \)), weighted F-measure (\({F_{\beta}^w}\)) and mean absolute error (\(MAE\)) to evaluate FOCUS. 

For SOD task, we follow \cite{Wang_2023_CVPR}, using DUTS-TR \cite{wang2017learning} as training dataset without extra data, evaluating our model on DUTS-TE, DUT-OMRON \cite{yang2013saliency}, HKU-IS \cite{li2015visual}, ECSSD \cite{shi2015hierarchical} and PACAL-S \cite{li2014secrets} respectively. We use \(\mathcal{S}_m\), \({E_{\xi}} \),  \(MAE\) as evaluation metrics for SOD. 

For SD, We use ISTD \cite{wang2018stacked} as our training and evaluation dataset and use balanced error rate (BCE) as the metric. For DBD, following previous work \cite{zhao2018defocus}, we use the combination of CUHK \cite{shi2014discriminative} and DUT \cite{zhao2018defocus} as training dataset, and the remaining 100 images in CUHK and 500 images in DUT for testing, and we use F-measure (\({F_{\beta}}\)) and \(MAE\) as metrics. Following \cite{wang2022objectformer}, we use CASIA-2.0 \cite{dong2013casia} as the training dataset and evaluate on CASIA-1.0, using pixel-level \(F1\) score and area under the curve (\(AUC\)) as evaluation metric.

\subsection{Implementation Details}

We use batch size 8 for all experiments and 2 NVIDIA A6000 GPUs with 48G memory. The FOCUS is trained on each training dataset with the size of \(512\times512\) for 20,000 iterations on average with AdamW optimizer \cite{loshchilov2017decoupled}. The initial learning rate is set to \(10^{-5}\) with a weight decay of 0.05 to regularize the model. The L2 norm is used for gradient clipping, and the maximum allowed value for gradients is set to 0.01. We use DINOv2-G \cite{oquab2023dinov2} pre-trained on ADE20K \cite{zhou2017scene} as the backbone for our SoTA model. Our framework is implemented using PyTorch 2.1.1 \cite{paszke2019pytorch}.


\begin{table*}[t]
\centering
\setlength{\tabcolsep}{1.5mm}
\scriptsize

\begin{tabular}{c c c c c c c c c c c c c c c c c c}
\toprule
\multicolumn{1}{c}{\multirow{2}{*}{\centering \textbf{id}}} & \multicolumn{1}{c}{\multirow{2}{*}{\centering \textbf{Variants}}} & \multicolumn{1}{c}{\multirow{2}{*}{\centering \textbf{Backbone}}}  & \multicolumn{1}{c}{\multirow{2}{*}{\centering \textbf{Trainable Param.}}} & \multicolumn{4}{c}{\textbf{Module/Method}} & \multicolumn{4}{c}{\textbf{COD}} & \multicolumn{4}{c}{\textbf{SOD}}\\
\cmidrule(lr){5-8} \cmidrule(lr){9-12} \cmidrule(lr){13-16} 
 &   & & & JP & CR & EE &PR  & \(\mathcal{S}_m \uparrow\)  & \({E_{\xi}} \uparrow\) & \({F_{\beta}^w} \uparrow\)  &  \(MAE \downarrow\) & \(\mathcal{S}_m \uparrow\)  & \({E_{\xi}} \uparrow\) & \({F_{\beta}} \uparrow\)  &  \(MAE \downarrow\)\\
\midrule

\textbf{0} &\textbf{Baseline} & DINOv2-L & 0.3G &  & &  & &.853&.931&.825&.043&.851&.919&.849&.054\\
\textbf{1} &\textbf{FOCUS} &DINOv2-L & 0.3G &\checkmark & & &  &.854&.931&.827&.042 &.853&.923&.847&.054 \\
\textbf{2} &\textbf{FOCUS} &DINOv2-L & 0.3G &\checkmark  &\checkmark & & &.861&.938&.836&.041&.855 &.926 &.851 &.052\\
\textbf{3} &\textbf{FOCUS} &DINOv2-L & 0.3G &\checkmark  &\checkmark &\checkmark & &.872&.937&.848&.041 &.870 &.922 &.864 &.051\\
\textbf{4} &\textbf{FOCUS} &DINOv2-G $\diamond$ & 0.1G &\checkmark  &\checkmark &\checkmark&  &.905 &.956 &.897 &.027  &.893	&.936	&.889 &.039\\
\textbf{5} &\textbf{FOCUS} &DINOv2-G & 1.2G &\checkmark  &\checkmark &\checkmark&  &.909 &.962 &.901& .026 &.897 &.942 &.896 &.037\\

\rowcolor{lightgray}
\textbf{6}&\textbf{FOCUS} & DINOv2-G & 1.2G &\checkmark  & \checkmark & \checkmark &\checkmark   &\textbf{.909}  &\textbf{.963}  &\textbf{.903} &\textbf{.025} &\textbf{.898}&\textbf{.943} &\textbf{.894} &\textbf{.037}\\
 

\bottomrule
\end{tabular}
\caption{Ablation study results of the proposed modules or methods of FOCUS, including CLIP Refiner (CR), Jointly Prediction (JP), Edge Enhancer (EE), and Pretrain (PR). $\diamond$ means training with the DINOv2 backbone frozen.}
\label{tab:ablation}
\end{table*}

\subsection{Main Results}
\subsubsection{Comparison of the state-of-the-art task-specific methods.}We compare our proposed FOCUS with recent models for COD including SINet \cite{fan2020camouflaged} , PFNet \cite{mei2021camouflaged} , ZoomNet \cite{pang2022zoom} , BSA-Net \cite{zhu2022can} , FSPNet \cite{huang2023feature} , ZoomNeXt \cite{pang2024zoomnext} and BiRefNet \cite{zheng2024bilateral} , models for SOD including MENet \cite{Wang_2023_CVPR} , SelfReformer \cite{10287608} , BBRF \cite{ma2021receptive} , and VST \cite{liu2021visual} , models for SD task including BDRAR \cite{zhu2018bidirectional} , DSD \cite{zheng2019distraction} , MTMT \cite{chen2020multi}, FDRNet \cite{zhu2021mitigating} and SILT \cite{yang2023silt}, models for DBD including DeFusionNet \cite{tang2020defusionnet} , CENet \cite{zhao2019enhancing} , DAD \cite{zhao2021self} , EFENet \cite{zhao2021defocus}  and DD \cite{cun2020defocus} , and models for FD including ManTra \cite{wu2019mantra} , SPAN \cite{hu2020span} , PSCCNet \cite{liu2022pscc} , TransForensics \cite{hao2021transforensics}  and ObjectFormer \cite{wang2022objectformer}. FOCUS outperforms these SoTA models on most metrics across 13 datasets covering 5 tasks. Table.~\ref{tab:COD}-\ref{tab:three_tables} shows quantitative comparisons between our proposed FOCUS with the previous SoTA models. Qualitative comparisons are in Fig .~\ref{qualitative1}.

In the most challenging foreground segmentation task, COD, which requires the model to recognize the object blending in its surroundings, FOCUS outperforms the existing  SoTA methods on most metrics across four mainstream datasets. For SOD tasks, FOCUS exceeds the task-specific models on almost all metrics, especially increasing in terms of \({E_{\xi}} \) by an average of 1.8\%. In SD tasks, FOCUS dramatically outperforms the previous SoTA on the ISTD dataset, with a 10.3\% decrease in BER. In the DBD task, FOCUS surpasses the previous SoTA by a 2.1\% increase on \({F_{\beta}}\) on DUT. In FD tasks, FOCUS also significantly surpasses previous SoTA models, with a 23.8\% increase on \(F1\) and a 3.8\% increase on \(AUC\). 


\subsubsection{Comparison of the state-of-the-art unified methods.}As previously mentioned, there is a lack of unified architecture to handle all foreground tasks. To the best of our knowledge, EVPv1 and EVPv2 \cite{liu2023explicit} are the most comparable works to our FOCUS in unifying foreground tasks. To demonstrate the superiority of FOCUS as a unified framework, we conducted extensive experiments comparing it with EVPv1 and EVPv2 across multiple datasets. Our results show that FOCUS consistently outperforms EVPv1 and EVPv2 in all metrics. This highlights the capability of FOCUS to handle a wide range of foreground segmentation tasks effectively, proving it can serve as a versatile and powerful model compared to existing unified methods.

\subsection{Ablation Study}

\begin{figure}[t]
\centering
\includegraphics[width=\columnwidth]{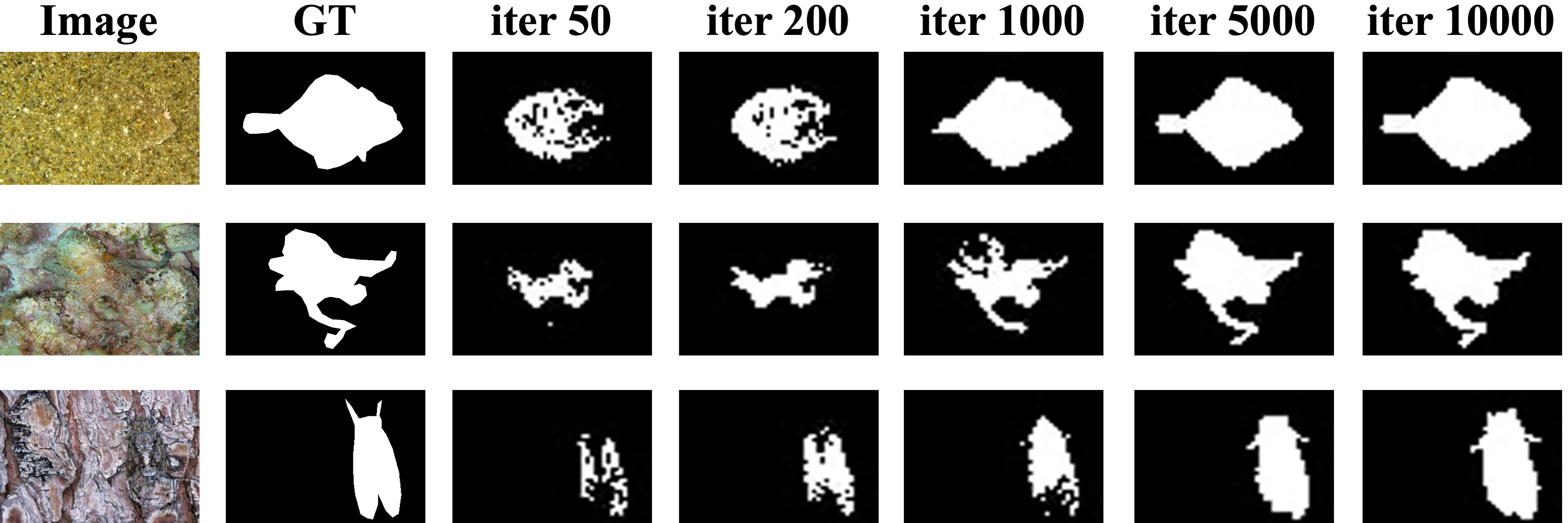} %
\caption{The visualization of PCA-based dimensionality reduction on the feature maps across different iterations. }
\label{pca}
\end{figure}

In this section, we conduct ablation experiments to analyze the properties of FOCUS. We use Mask2Former equipped with the DINOv2-L backbone as a robust baseline and choose the most representative foreground segmentation tasks, COD and SOD, as the ablation tasks. We select the mainstream dataset CAMO and PASCAL-S for COD and SOD respectively. To ensure consistency, all experiments were conducted using the same training recipe, with a batch size of 2. The training iterations are set to 10,000 for COD and 20,000 for SOD. Quantitative results related to each module or method are shown in Table.~\ref{tab:ablation}.

As shown in the table, variants of FOCUS with the CLIP refiner perform better than those without it, thanks to the multi-modal knowledge distilled from CLIP. We set the variants with joint prediction to perform foreground segmentation and background segmentation jointly, the comparison with the baseline shows that it can slightly improve the performance of FOCUS. Additionally, with the help of the edge enhancer to inject edge information of the object into the backbone image feature, the performance of variants of DINOv2 significantly improves in the provided metrics. We also evaluate the effectiveness of pretraining on ADE20K, which demonstrates modest improvements. 

We use DINOv2-G as the backbone for our SoTA models, which inevitably results in a large number of parameters. To ensure a fair comparison, we freeze the DINOv2-G backbone, limiting the number of trainable parameters in our model to 0.1G. The results indicate a slight decrease in performance compared to the fully fine-tuned version. However, when compared to models like BiRefNet (215M) and SelfReformer (\textasciitilde220M), the frozen-backbone FOCUS still matches or surpasses previous state-of-the-art performance, despite having fewer trainable parameters.

We initialize the first layer of the transformer decoder with PCA-reduced feature maps from the backbone in our paper. As shown in Fig.~\ref{pca}, these PCA-reduced feature maps begin to exhibit strong semantic features in the early training stages. As training progresses, we are pleasantly surprised to find that even without further forward propagation, the patch-level feature maps, simply reduced by PCA, are able to approach the ground truth quality. Using them for initialization, compared to random initialization, provides a valuable spatial prior for subsequent mask attention.

\section{Conclusion}

In this paper, we propose FOCUS, a unified multi-modal approach to tackle multiple subdivision tasks of foreground segmentation. We leverage the concept of object queries to handle foreground segmentation tasks and develop a multi-scale semantic network that simultaneously performs foreground and background segmentation, fully utilizing the background information of the image to optimize prediction. We also introduced a novel distillation method integrating the contrastive learning strategy to enhance boundary-aware foreground segmentation. Theoretically, our model can be extended to any foreground segmentation task. Extensive experiments conducted on diverse datasets demonstrate the effectiveness of our proposed framework.

\section{Acknowledgments}
This project was supported by NSFC under Grant No.  62102092.

\bibliography{aaai25}

\end{document}